\documentclass[a4paper,twoside]{article}

\usepackage{epsfig}
\usepackage{subcaption}
\usepackage{calc}
\usepackage{amssymb}
\usepackage{amstext}
\usepackage{amsmath}
\usepackage{amsthm}
\usepackage{multicol}
\usepackage{pslatex}
\usepackage{apalike}
\usepackage{SCITEPRESS} 
\usepackage{graphicx}

\begin{document}

\title{It Means More if It Sounds Good: \\ Yet Another Hypothesis Concerning the Evolution of Polysemous Words}

\author{\authorname{Ivan P. Yamshchikov\sup{1}\orcidAuthor{0000-0003-3784-0671}, Cyrille Merleau Nono Saha\sup{1}, Igor Samenko\sup{2}, J{\"u}rgen Jost\sup{1}}
\affiliation{\sup{1}Max Planck Institute for Mathematics in the Sciences, Inselstrasse 22, Leipzig, Germany \\
\sup{2} Institute of Computational Technologies SB RAS}
\email{ivan@yamshchikov.info, saha@mis.mpg.de, i.samenko@gmail.com, jjost@mis.mpg.de}
}


\abstract{This position paper looks into the formation of language and shows ties between structural properties of the words in the English language and their polysemy. Using Ollivier-Ricci curvature over a large graph of synonyms to estimate polysemy it shows empirically that the words that arguably are easier to pronounce also tend to have multiple meanings.}

\onecolumn \maketitle \normalsize \setcounter{footnote}{0} \vfill

\section{\uppercase{Introduction}}
\label{sec:introduction}

\noindent Starting form the second half of the nineteenth century \cite{schleicher1869darwinism} various researchers address historic development of language from evolutionary grounds. 

A considerable proportion of the works in this field use word frequency as an important proxy of the word fitness. For example, \cite{pagel2007frequency} demonstrate across several languages that frequently used words evolve at slower rates, whereas infrequently used words evolve more rapidly. \cite{newberry2017detecting} state that a possible explanation for this phenomenon could be a stronger stochastic drift of rare words. In the meantime, \cite{adelman2006contextual} notice that word frequency is confounded with polysemy, i.e., the number of contexts in which a word has been seen. They also show that this contextual diversity is a crucial factor that determines word-naming and lexical decision times. 

\cite{lee1990some} demonstrates that older words are more polysemous than recent words and that frequently used words are more polysemous than infrequently used words. This goes in line with \cite{maccormac1985cognitive} theory of semantic conceptual change that states that words evolve additional meanings through metaphor. It seems that the frequency of the word is confounded with its semantics. 

\cite{bybee2002word} reviews results on how a sound change affects the lexicon and documents that a sound change affects high-frequency words and low-frequency words differently. This shows that the frequency of the word is confounded with its phonetic properties. The ideas that there is a subtle correspondence between phonetics and semantics were entertained by literary theorists \cite{shklovsky1917art} and artists  \cite{Kruchenykh} at least from the beginning of the twentieth century. In a massive study across nearly two-thirds of the world's languages \cite{Blasi} managed to demonstrate that a considerable proportion of 100 essential vocabulary items carry strong associations with specific kinds of human speech sounds, occurring persistently across continents and linguistic lineages. \cite{DBS} showed that modern methods of computational linguistics could be used to highlight such associative structures within a language.

This position paper develops these ideas, and states that the phonetic simplicity of a word is to some extent correlated with the number of its semantic contexts. We also speculate on possible cognitive mechanisms underlying this connection. 

\section{\uppercase{Data}}
\label{sec:data}

\noindent To relate polysemous properties of the words with their phonetic structure, we use two different datasets represented as graphs with words as vertices.

\cite{smerlak2020localization} reconsidered Maynard Smith’s toy model of protein evolution \cite{smith1970natural} in context of {\em neutral evolution}. We use this dataset here for a different purpose. Let us consider a set of all possible four-letter words and assume that two words are connected with an edge if the second word can be derived from the first one with an edit of one letter. We further call this graph a {\em graph of edits}. One could interpret the degree of the vertices in this graph as a proxy of the word's phonetic simplicity. Indeed, if a word has more one-letter edits that produce a meaningful word, one can assume that it consists of the letters or sounds with higher joined probabilities. We discuss this hypothesis further in Section \ref{sec:PP}.

\begin{figure*}[h!]
 \includegraphics[width=\textwidth]{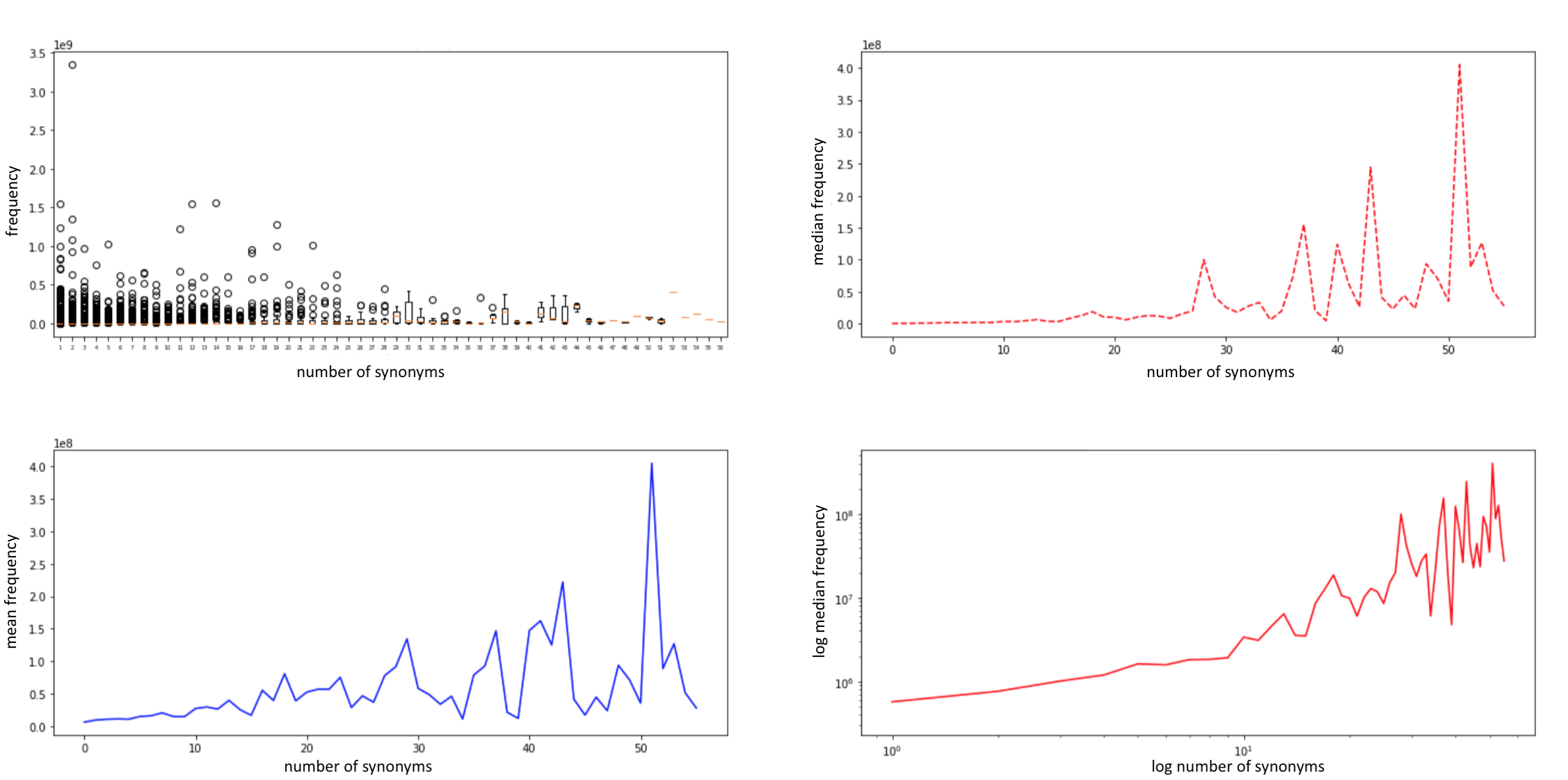}
\caption{
Word frequency tends to be higher for the words that have more synonyms.}
\label{fig:over}    
\end{figure*}

We provide\footnote{https://github.com/i-samenko/Triplet-net/tree/master/data} a large dataset of English synonyms that is based on WordNet\footnote{https://wordnet.princeton.edu/}. Here two words are connected with an edge if they are synonymous. Figure \ref{fig:over} shows the frequency of the word usage (estimated on a large chunk of English Wikipedia\footnote{https://www.kaggle.com/rtatman/english-word-frequency}) as a function of the number of synonyms that it has in the graph. This is a well-known fact that is important for us here, since to a certain extent it validates the dataset of synonyms as a representative one.  

\begin{figure}[h!]
 \includegraphics[width=0.45\textwidth]{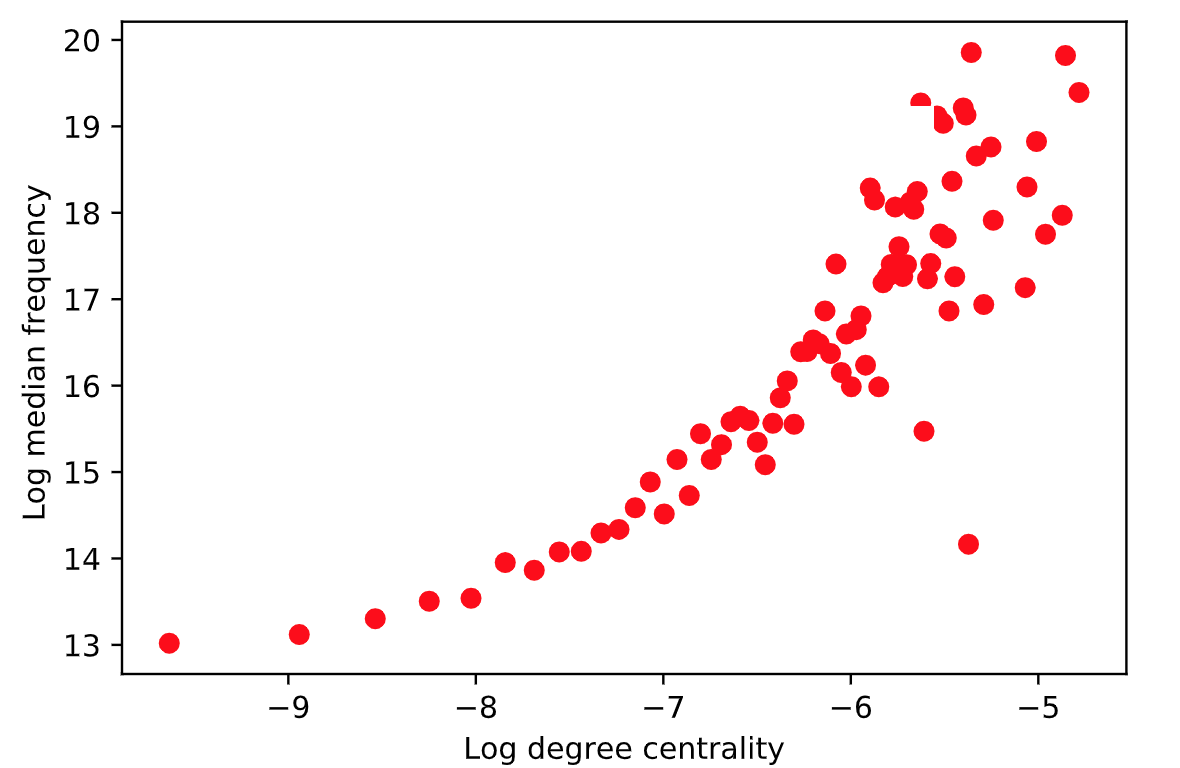}
\caption{
Log of degree centrality in the graph of synonyms and frequency of use for the four-letter words that form the graph of edits.}
\label{fig:fc}    
\end{figure}

Figure \ref{fig:fc} shows that the connection between word frequencies and degrees in the graph of synonyms is even stronger for the four-letter words that form the graph of edits.

The sheer number of synonyms adjacent to a given word does not necessarily correspond to the number of various semantic contexts in which it can occur. It is well known that polysemic words have more synonyms and tend to have higher frequencies, but one can not infer the number of possible semantic contexts in which a word can occur our of the number of its synonyms. Further, we discuss how one can estimate polysemy of a word using the geometry of the graph of synonyms. 

\section{\uppercase{Ollivier-Ricci Curvature and Polysemy}}
\label{sec:ORP}

\noindent Ollivier-Ricci curvature \cite{ollivier2009ricci} is commonly used for community detection \cite{ni2015ricci}, \cite{sia2019ollivier}.  In this paper we use it in a way that is novel for mathematical linguistic and claim that it could be used as a proxy for the word's polysemy measure in the language. Yet before we get into the details let us briefly describe Ollivier-Ricci curvature itself.

Here we use the method that \cite{ni2019community} provide for calculation of the curvature\footnote{https://github.com/saibalmars/GraphRicciCurvature}. One considers a particular probability distribution $m_x$, which has parameter $\alpha$, and a graph $G$. For a vertex
$x \in G$ with degree $k$, let $\Gamma(x) = \{x_1, x_2, ..., x_k\}$ denote the set of neighbors of $x$. For any $\alpha \in [0,1]$ the probability measure $m^{\alpha}_x$ is defined as

\begin{equation}
m^{\alpha}_x (x_i) = \begin{cases}
\alpha \hspace{3pt} \text{if} \hspace{3pt} x_i = x \\
(1 - \alpha)/k \hspace{3pt} \text{if} \hspace{3pt} x_i \in \Gamma(x)\\
0 \hspace{3pt} \text{otherwise}\\

\end{cases}
\end{equation}

\begin{figure}[h!]
 \includegraphics[width=0.45\textwidth]{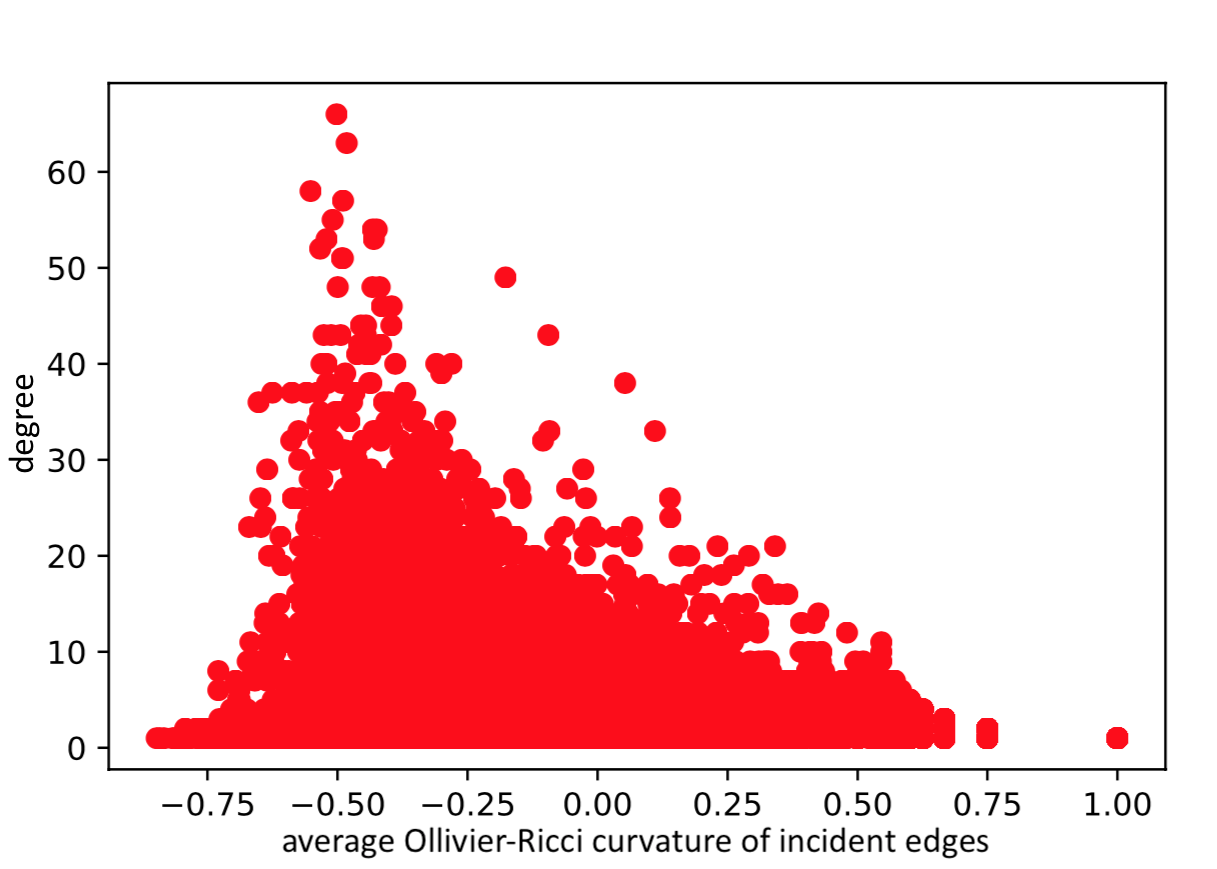}
\caption{
Words with lower average Ollivier-Ricci curvature of incident edges tend to have higher degree in the graph of synonyms.}
\label{fig:points}    
\end{figure}

The intuition behind the curvature of a given edge in our case is rather intuitive. Once an edge is within a dense community it has positive curvature, whereas edges that connect separate communities have negative curvature. This property of Ollivier-Ricci curvature directly leads to the detection of polysemy of a given word. Indeed, every incident edge with a positive Ollivier-Ricci curvature would connect the word to a synonym within the same semantic context, however, an incident edge with negative Ollivier-Ricci curvature points to a synonym within a drastically different semantic field. Therefore, one can use the number of incident edges with negative Ollivier-Ricci curvature or the average Ollivier-Ricci curvature across incident edges as a measure of the polysemy of the word. Figure \ref{fig:points} shows that words with lower average Ollivier-Ricci curvature of incident edges tend to have a higher degree in the graph of synonyms. This also goes in line with the statement that word frequency is confounded with polysemy. 

Further, we show that the situation is more nuanced and that there is a connection between the location of the word within the graph of edits and its polysemy.

\section{\uppercase{Polysemy and Phonetics}}
\label{sec:PP}

\begin{figure}[h!]
 \includegraphics[width=0.45\textwidth]{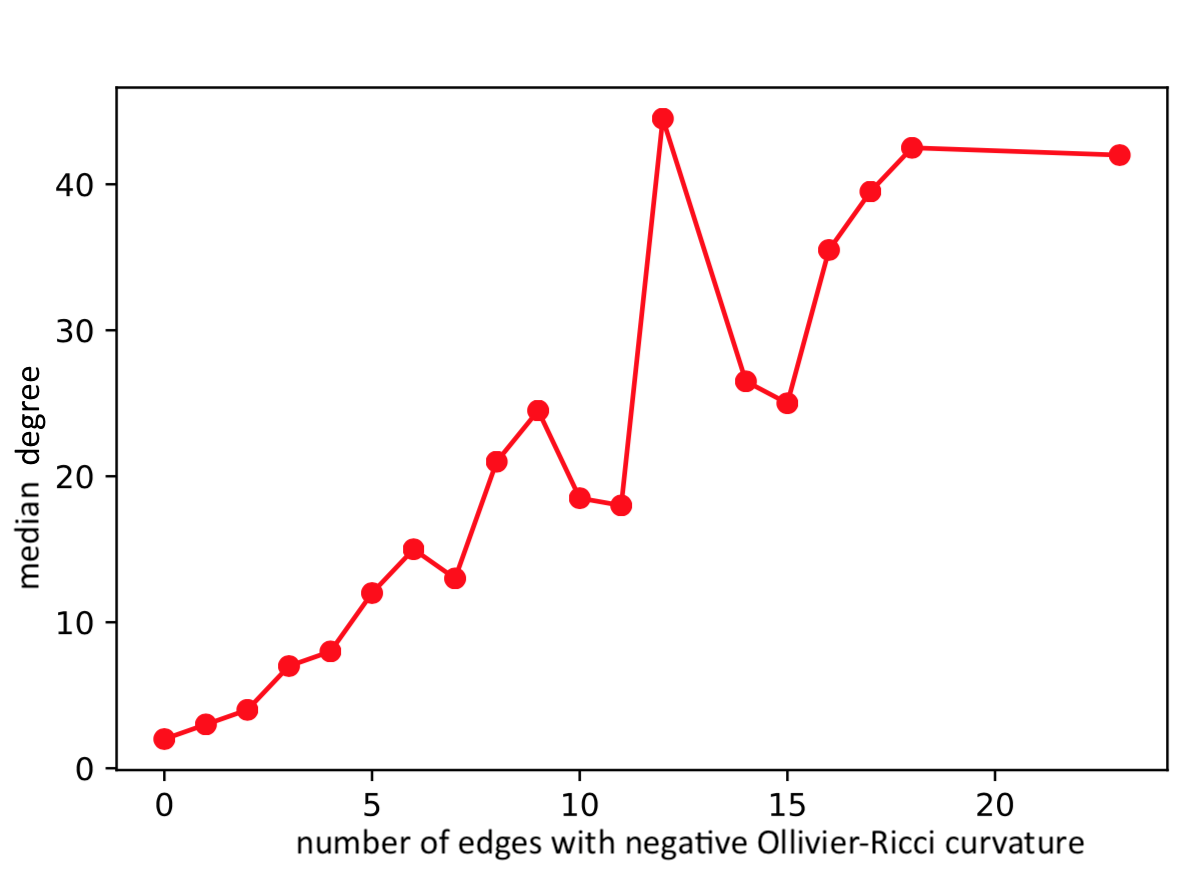}
\caption{
Median degree of the words in the synonym graph as a function of the number of incident edges with negative Ollivier-Ricci curvature. The words with more synonyms tend to be more polysemous.}
\label{fig:1}    
\end{figure}

\noindent Figure \ref{fig:1} shows how the degree of a word in the graph of synonyms depends on its polysemy, i.e., the number of incident edges with negative Ollivier-Ricci curvature in the graph of synonyms. This connection is well known and can be seen in the proposed dataset.

\begin{figure}[h!]
 \includegraphics[width=0.45\textwidth]{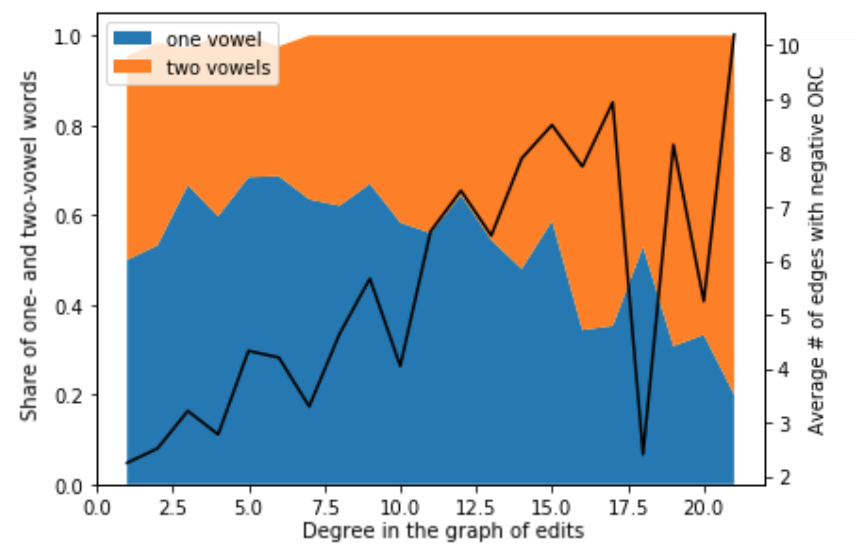}
\caption{
The words with a higher degree in the graph of edits tend to have two vowels rather than one. This words also tend to be more polysemous, since a higher number of incident edges with negative Ollivier-Ricci curvature could be associated with more separate semantic contexts in which a word could be used.}
\label{fig:2}    
\end{figure}

Let us now discuss the graph of edits. One can regard the formation of actual words as a purely random process. The \cite{smith1970natural} toy-model is based on the assumption that if we regard all possible one-letter edits of a word, any combination of letters is equally 'lucky' to become be another meaningful word. However, \cite{nowak1999evolution} show that introduction of an error in sound recognition on the stage of a protolanguage makes it very limited: "Adding new sounds increases the number
of objects that can be described but at the cost of an increased
probability of making mistakes; the overall ability to transfer
information does not improve". The authors show that combining sounds into words is a way to overcome such error limit. In line with this reasoning we suggest to look at the graph of edits from a phonetic perspective. Indeed, one should remember that there are certain phonetic structures that are more characteristic for a given language. Moreover, if a combination of letters is 'not pronounceable' it definitely can not be a meaningful word. Therefore, one can suppose that the degree in the graph of edits corresponds to the so-called 'phonetic simplicity' of a word. The words that are easier to pronounce would probably have a higher degree in the graph edits. Figure \ref{fig:2} partially illustrates this supposition.

\begin{figure}[h!]
 \includegraphics[width=0.45\textwidth]{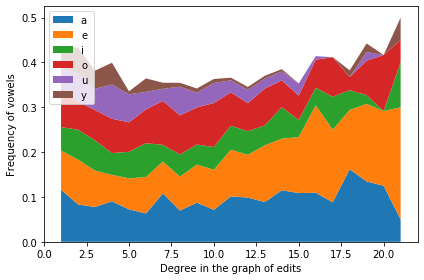}
\caption{Relative frequencies of vowels are different for words with different degrees in the graph of edits. "i" and "u" tend to occur less often in highly connected words, whereas "e" and "a" are more common.
}
\label{fig:dist}    
\end{figure}

Figure \ref{fig:2} and Figure \ref{fig:dist} show that as the degree of the words in the graph of edits gets higher the words tend to have two vowels rather than one. Also "u" and "i" are less frequent among densely connected words, however "a" and "e" seem to occur more often. "y" already vanishes as the degree of the words gets bigger than ten.

Table \ref{tab:dataform} shows that the frequency of two-vowel words that are arguably more robust in terms of phonetic simplicity correlates with the degree of the corresponding word in the graph of synonyms. It also correlates with the number of incident edges with negative Ollivier-Ricci curvature in the graph of synonyms. Finally, there is a strong correlation between the frequency of the two-vowel words and their degree in the graph of edits. 

\begin{table}
\centering
\small{\begin{tabular}{ll}
 \hline
Value & correlation with frequency \\
 & of two-vowel words\\
\hline
\# of incident edges & $59.5$\%\\
with negative ORC & \\
\hline
degree in the graph & $57.9$\%\\
of synonyms & \\
 \hline
degree in the graph & $74.7$\%\\
of edits & \\
\hline
\end{tabular}}
\caption{Both number of incident edges with negative Ollivier-Ricci curvature and degree in the graph of synonyms correlate with frequencies of two-vowel words in the graph of edits. Degree in the graph of edits shows even stronger correlation with the frequency of the two-vowel words.}
  \label{tab:dataform}
\end{table}

All these observed correlations allow speculating that the structure of the graph of edits is affected by certain phonetic properties of the English language. A higher degree of a word in this graph seems to capture certain phonetic usability of this word. 

\section{\uppercase{Discussion}}
\label{sec:discussion}

\noindent This position paper demonstrates an interesting empirical fact: there is a connection between the structure of the graph of edits that is based on purely formal reasoning and a graph of synonyms that to a certain extent captures semantic complexity of the language. This fact in itself is thought-provoking. It motivates a search for a phonetically inspired notion of fitness that could be applied to the problems of the evolution of language. However, the discussion of such a notion is out of the scope of this work. Here we would only like to highlight the role of negatively curved incident edges in the graph of synonyms. We hope that this geometric approach could be further used to study polysemy. 

Let us now briefly discuss the final interesting connection between the phonetic structure of the words and their polysemy. Out of Table \ref{tab:dataform} we know that the correlation between the degree in the graph of edits and the frequency of two vowel words is above 74\%. We also know that the frequency of two-vowel words correlates with a number of incident negatively curved edges in the graph of synonyms and with the degree of the word in the graph of synonyms. These two quantities are also strongly correlated. In fact, the degree of the word in the graph of synonyms and the number of its incident negatively curved edges correlate with a coefficient of 0.97. Indeed, a number of synonyms, polysemy, and frequency of use are known to be correlated. However, we would like to discuss another interesting empiric connection here that could highlight the connection between these properties of the words and their phonetics.

\begin{figure}[h!]
 \includegraphics[width=0.45\textwidth]{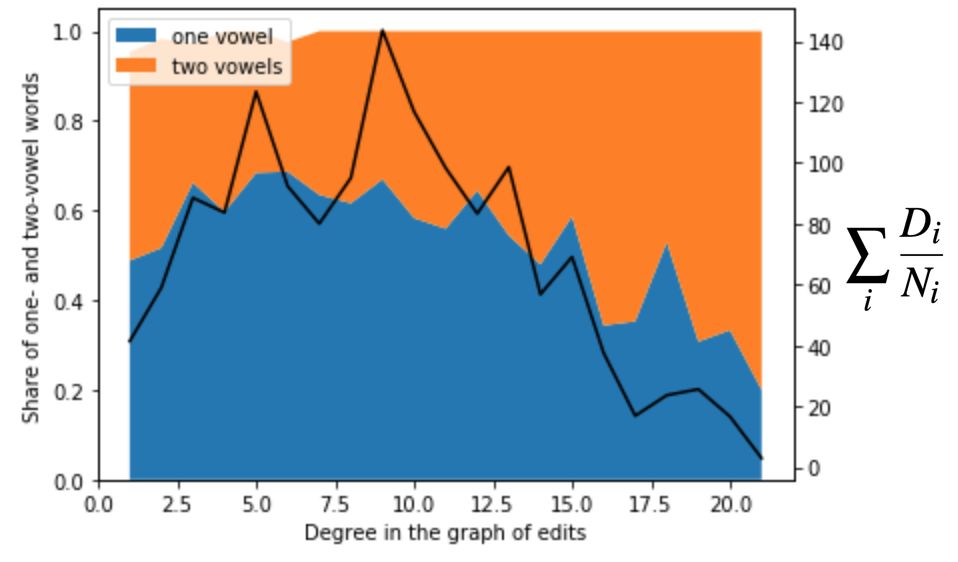}
\caption{
Sum of the ratios between degree and number of incident edges with negative Ollivier-Ricci curvature across all words with a fixed degree in the graph of edits correlates with the frequency of two-vowel words.}
\label{fig:vow}    
\end{figure}

Let us regard all the words with a given degree in the graph of edits. For a given word $i$ let us count all incident edgers with negative Olliver-Ricci curvature in the synonym graph and let us denote this number as $N_i$. Let us also denote the degree of this word in the graph of synonyms as $D_i$. Let us then calculate the ration $\frac{D_i}{N_i}$. Figure \ref{fig:vow} demonstrates how the sum of these ratios across all words with a fixed degree in the graph of edits $\sum_i \frac{D_i}{N_i}$ depends on this degree.

Metric in Figure \ref{fig:vow} correlates with the frequency of two vowel words with -83.4\%. In our opinion, this might highlight the importance of Ollivier-Ricci curvature-based polysemy measure as a tool to highlight the connection between polysemy and phonetic properties of the words. It stands to reason that the words that are easier to pronounce would be used more often and acquire more synonyms with time. This highlights the possibility that polysemy could be associated with certain acoustic simplicity. Therefore it develops the idea of evolution through metaphor stated in \cite{maccormac1985cognitive}, showing that the words that are easier to pronounce could be more prone to such evolution and, as time proceeds, could end up with more semantic fields.

\section{\uppercase{Conclusion}}

This position paper demonstrates empirically a connection between polysemy of the words and their formal structure. We propose to use Ollivier-Ricci curvature over a graph of synonyms as an estimate for polysemy of the word. We speculate that the aforementioned connection between polysemy and formal structure is rooted in the phonetic properties of the language. We empirically demonstrate that certain phonetic properties of the words are correlated with their polysemy.

\vfill
\section*{\uppercase{Acknowledgements}}
\noindent Authors are extremely grateful to Matteo Smerlak and Massimo Warglien for the help, support and constructive discussions.

\noindent

\bibliographystyle{apalike}
{\small
\bibliography{example}}

\end{document}